\documentclass[10pt,twocolumn,letterpaper]{article}

\usepackage{cvpr}
\usepackage{times}
\usepackage{epsfig}
\usepackage{graphicx}
\usepackage{amsmath}
\usepackage{amssymb}
\usepackage{caption}

\usepackage{multirow}
\usepackage{adjustbox}

\usepackage[pagebackref=true,breaklinks=true,letterpaper=true,colorlinks,bookmarks=false]{hyperref}

\cvprfinalcopy % *** Uncomment this line for the final submission

\def\cvprPaperID{6} % *** Enter the CVPR Paper ID here

\newcommand\blfootnote[1]{%
  \begingroup
  \renewcommand\thefootnote{}\footnote{#1}%
  \addtocounter{footnote}{-1}%
  \endgroup
}

\ifcvprfinal\fi
\begin{document}

%%%%%%%%% TITLE
\title{An Effective Pipeline for a Real-world Clothes Retrieval System}

\author{Yang-Ho Ji$^1$\footnotemark[1] \footnotemark[2],\space\space
HeeJae Jun$^2$\footnotemark[1] \footnotemark[2],\space\space
Insik Kim$^3$\footnotemark[1] \footnotemark[2],\space\space
Jongtack Kim$^4$\footnotemark[1] \footnotemark[2],\space\space
Youngjoon Kim$^5$\footnotemark[1] \footnotemark[2],\\
Byungsoo Ko$^6$\footnotemark[1] \footnotemark[2],\space\space
Hyong-Keun Kook$^7$\footnotemark[1] \footnotemark[3],\space\space
Jingeun Lee$^8$\footnotemark[1] \footnotemark[2],\space\space
Sangwon Lee$^9$\footnotemark[1] \footnotemark[2],\space\space
Sanghyuk Park$^{10}$\footnotemark[1] \footnotemark[2]\\
NAVER/LINE Vision\footnotemark[2],\space\space Search Solution Vision\footnotemark[3]\\
{\tt\small \{arnilone$^1$, kobiso62$^6$, hyongkuen63$^7$, jglee0206$^8$, shine0624$^{10}$\}@gmail.com} \\
{\tt\small \{heejae.jun$^2$, insik.kim$^3$, jongtack.kim$^4$, kim.youngjoon$^5$, leee.sangwon$^9$\}@navercorp.com}
}

% \begin{document}
\twocolumn[{%
\renewcommand\twocolumn[1][]{#1}%
\maketitle
\begin{center}
  \centering
  \includegraphics[width=\textwidth]{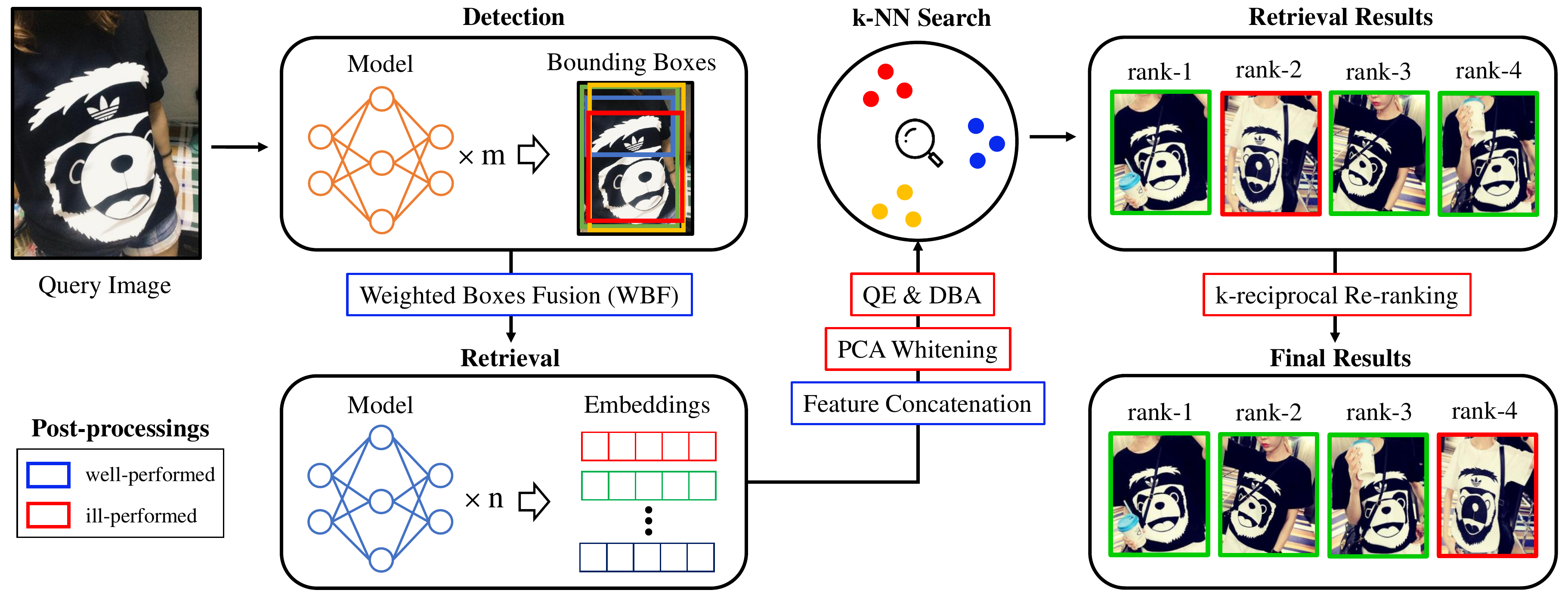}
  \captionof{figure}{The proposed pipeline for real-world clothes retrieval that consists of three components: detection, retrieval and post-processing. Blue boxes indicate the post-processing methods that performed well while red is for those that did not. }
  \label{fig:short}
  \vspace{-0.5mm}
\end{center}%
}]

\blfootnote{*All authors from Visual Search \& Recommendation team at NAVER/LINE corporation contributed equally to this work and are listed in alphabetical order by last name.}

%%%%%%%%% ABSTRACT
\begin{abstract}
In this paper, we propose an effective pipeline for clothes retrieval system which has sturdiness on large-scale real-world fashion data. Our proposed method consists of three components: detection, retrieval, and post-processing. We firstly conduct a detection task for precise retrieval on target clothes, then retrieve the corresponding items with the metric learning-based model. To improve the retrieval robustness against noise and misleading bounding boxes, we apply post-processing methods such as weighted boxes fusion and feature concatenation. With the proposed methodology, we achieved 2nd place in the DeepFashion2 Clothes Retrieval 2020 challenge.

\end{abstract}

%-------------------------------------------------------------------------
%%%%%%%%% BODY TEXT
\section{Introduction}
Recently, fashion domain has attracted much attention in computer vision research. Similarly, demand for online shopping for `fashionable' items has also emerged to be one of the world's largest industries. Thus, it has become crucial for advanced visual clothing retrieval systems to provide the finest shopping experience for online customers.

In this paper, we present an effective pipeline for a real-world clothes retrieval system using large-scale fashion data. The clothes retrieval system includes two vision tasks: detection and retrieval. Since the advent of deep convolutional neural networks (DCNN), vision tasks have improved to show astonishing results thus we employ modern variants of these DCNN models in our detection and retrieval network. In addition, we take advantage of popular post-processing methods to further improve the performance of the retrieval system: weighted boxes fusion (WBF) methods to effectively combine the predicted boxes of multiple detectors, and feature concatenation to elevate the similarities of the feature embeddings from different DCNN models.

%-------------------------------------------------------------------------
\section{Methodology}
In this section we present our method of clothes retrieval which consists of three components: detection, retrieval, and post-processing. Figure~\ref{fig:short} depicts the whole pipeline.

\begin{table} % [!ht]
\begin{center}
\begin{adjustbox}{width=0.75\columnwidth,center}
\begin{tabular}{|l|c|c|}
\hline
Dataset & \# of pairs & \# of images \\
\hline\hline
DeepFashion2~\cite{DeepFashion2} & 873,234 & 491,895 \\
\hline
DARN~\cite{Huang_2015_ICCV} & 91,390 & 182,780 \\
\hline
Street2Shop~\cite{WhereToBuyItICCV15} & 39,479 & 416,840 \\
\hline
Zalando~\cite{han2017viton} & 16,253 & 32,642 \\
\hline
MVC~\cite{Liu2016MVCAD} & 36,656 & 156,449 \\
\hline
MPV~\cite{Dong_2019_ICCV} & 13,524 & 50,290 \\
\hline
\end{tabular}
\end{adjustbox}
\end{center}
\vspace{-6mm}
\caption{The utilized fashion-related datasets.}
% \vspace{-5mm}
\label{table:dataset}
\vspace{-3mm}
\end{table}

\begin{table}
\begin{center}
\begin{adjustbox}{width=1.0\columnwidth,center}
\begin{tabular}{|l|c c c|c c c|}
\hline
Model & $AP$ & $AP_{50}$ & $AP_{75}$ & $AP_{S}$ & $AP_{M}$ & $AP_{L}$ \\
\hline\hline
ATSS & 72.8 & 83.6 & 80.6 & 52.7 & 53.2 & 73.0 \\
Cascade Mask R-CNN & 72.7 & 83.9 & 80.4 & 42.6 & 52.0 & 68.4 \\
CenterNet & 70.4 & 81.4 & 77.9 & 30.1 & 44.3 & 70.8 \\
RetinaNet-R101-FPN & 67.4 & 81.1 & 76.2 & 32.7 & 46.1 & 67.6 \\
RetinaNet-X101-FPN & 67.4 & 81.2 & 76.1 & 32.7 & 51.9 & 67.6 \\
\hline
\end{tabular}
\end{adjustbox}
\end{center}
\vspace{-6mm}
\caption{Detection results on DeepFashion2 \textit{validation set}.}
\label{table:detection_result}
\vspace{-3mm}
\end{table}

\subsection{Detection}
To retrieve fashion items correctly on an image, it is necessary to detect each fashion item via a detection network. We leverage five state-of-the-art object detection models to get well-localized and classified clothing boxes: adaptive training sample selection (ATSS)~\cite{zhang2020bridging}, cascaded mask R-CNN~\cite{Cai_2018_CVPR}, CenterNet~\cite{1904.07850}, RetinaNet-R101-FPN, and RetinaNet-X101-FPN~\cite{Lin_2017_ICCV}. We report detection performances in detail on Section~\ref{sec:detection_results}.

\subsection{Retrieval}
To perform retrieval tasks based on the detected clothing items, we evaluate prevalent DCNN models (e.g. ResNet152~\cite{he2015deep} and SE-ResNeXt101~\cite{hu2017squeezeandexcitation}), optimize loss functions and fine-tune hyperparameters to decide on the well performing options. In addition, we leveraged a unique framework called combination of multiple global descriptors (CGD)~\cite{1903.10663} which has been proven to increase performance of a model by combining multiple global descriptors, such as maximum activation of convolutions (MAC)~\cite{tolias2015particular} and generalized mean pooling (GeM)~\cite{radenovi2017finetuning}. We report the performances of the utilized models in Section~\ref{sec:retrieval_results}.

\subsection{Post-processing}
To boost the performance of the clothes retrieval task, we exploit several post-processing methods. We create ensembles using detection and retrieval results as well as experiment with other widely-used techniques, such as feature manipulation and re-ranking. We report the effectiveness of these post-processing methods on Section~\ref{sec:post-processing_results}.

\vspace{2mm}
\noindent\textbf{Weighted Boxes Fusion}
In order to enhance the retrieval performance even further, we adopt the WBF~\cite{1910.13302} technique of creating well-cropped query and database images from the results of multiple detectors. WBF collates all box information from our detection network and provides more accurate predictions which we found to correct inconsistencies in results.

\vspace{2mm}
\noindent\textbf{Feature Concatenation}
As each model captures different features of a given image, we ultimately concatenate \textit{l2}-normalized feature embeddings to increase our accuracy when retrieving images. This method has a trade-off with computation time and memory being used; however, each addition of features helped to achieve better performance in the task of image retrieval.

%-------------------------------------------------------------------------
\section{Experiments}
In this section, we describe the datasets that were used and report experimental result details. Included are the performances of each component of our retrieval pipeline and discussions about the effectiveness of post-processing methods that were assessed.

\subsection{Dataset}
Table ~\ref{table:dataset} shows various fashion-related datasets that were used in detection and retrieval tasks. Specifically, original images were used to train detection models while cropped and deduplicated images were for the retrieval models.

\subsection{Detection Results}
\label{sec:detection_results}
We train our detection models on DeepFashion2 \textit{train set} only, then validate their performances on DeepFashion2 \textit{validation set}. On Table ~\ref{table:detection_result}, we summarize the evaluated detection performances on validation set. 

\begin{table}
\begin{center}
\begin{adjustbox}{width=1.0\columnwidth,center}
\begin{tabular}{|l|c|c|}
\hline
Model & Description & Acc@10 \\
\hline\hline
m0 & used DeepFashion2 train set only to train & 0.868 \\
m1 & +Street2Shop & 0.873 \\
m2 & +DARN+MVC & 0.873	 \\
m3 & +Street2Shop+DARN & 0.878 \\
m4 & +Street2Shop+DARN+MVC & 0.879 \\
m5 & +Street2Shop+DARN+MVC+MPV & 0.877 \\
m6 & +Street2Shop+DARN+MVC+MPV+Zalando & 0.878 \\
m7 & used DeepFashion2 train and valid set to train & - \\
\hline
m8 & used hparams by automl for msloss on m0 & 0.871 \\
m9 & changed backbone to SENext101 on m8 & 0.858 \\
m10 & increased resol. to 224x384 on m0 & 0.870 \\
m11 & used GeM+MAC descriptors on m4 & 0.870 \\
m12 & changed lr from 0.2 to 0.17 on m11 & 0.877 \\
m13 & changed lr from 0.2 to 0.17 on m4 & 0.878 \\
m14 & used hparams by automl for msloss on m13 & 0.881 \\
m15 & changed lr from 0.2 to 0.17 on m6 & \textbf{0.882} \\
m16 & used hparams by automl for msloss on m15 & 0.880 \\
m17 & used GeM+SPoC descriptors on m4 & 0.870 \\
\hline
\end{tabular}
\end{adjustbox}
\end{center}
\vspace{-6mm}
\caption{Top-10 accuracy of retrieval models with ground truth bounding boxes on DeepFashion2
\textit{validation set}.}
\label{table:exp_retrieval_model}
\vspace{-5mm}
\end{table}

\begin{table}
\begin{center}
\begin{adjustbox}{width=1.0\columnwidth,center}
\begin{tabular}{|l|c||l|c|}
\hline
Param         & Value             & Param                & Value     \\ \hline\hline
Backbone      & ResNet152-D~\cite{he2019bag}    & Descriptor    & GeM~\cite{radenovic2018fine}       \\ \hline
Input size  & 224$\times$224    & Output dim.     & 1,024     \\ \hline
Total epochs  & 200               & Optimizer            & SGD       \\ \hline
Init. LR & 0.2               & LR decay             & cosine    \\ \hline
Losses          & \multicolumn{3}{c|}{Softmax + Center~\cite{wen2016discriminative} + Triplet~\cite{schroff2015facenet} + MS~\cite{wang2019multi}} \\ \hline
\end{tabular}
\end{adjustbox}
\end{center}
\vspace{-5mm}
\caption{Default settings for training retrieval models.}
\label{table:setting}
\vspace{-2mm}
\end{table}

\begin{table}
\begin{center}
\begin{adjustbox}{width=1.0\columnwidth,center}
\begin{tabular}{|l|c c c c c|}
\hline
Models & ${WBF}_{1}$ & ${WBF}_{2}$ & ${WBF}_{3}$ & ${WBF}_{4}$ & ${WBF}_{5}$ \\
\hline\hline
ATSS & \checkmark & \checkmark & \checkmark & \checkmark & \checkmark   \\
+ Cascade Mask R-CNN & & \checkmark & \checkmark & \checkmark & \checkmark  \\
+ CenterNet & & & \checkmark & \checkmark & \checkmark \\
+ RetinaNet-R101-FPN & & & & \checkmark & \checkmark  \\
+ RetinaNet-X101-FPN & & & & & \checkmark  \\
\hline
$AP_{50}\,(\%)$ & 83.6 & 84.0 & 84.9 & 85.4 & \textbf{85.8}\\
\hline
\end{tabular}
\end{adjustbox}
\end{center}
\vspace{-5mm}
\caption{Weighted boxes fusion (WBF) results on DeepFashion2 \textit{validation set}. $WBF_{n}$ refers WBF result with $n$ models. There are performance gains as more detection models are added.}
\label{table:wbf_result}
\vspace{-3mm}
\end{table}

\subsection{Retrieval Results} \label{sec:retrieval_results}
Table ~\ref{table:exp_retrieval_model} shows the performances of retrieval models using ground-truth bounding boxes on the DeepFashion2 validation set. In addition to the DeepFashion2 \textit{train set}, we accumulated publicly available datasets as listed in Table~\ref{table:dataset} to train our retrieval models. The default configurations used to train retrieval models is described in the Table~\ref{table:setting}.

\subsection{Post-processing Results}
\label{sec:post-processing_results}
We summarize the aforementioned post-processing methods with a description of how it was applied and whether it worked to improve image retrieval result.

\subsubsection{What worked well}

\vspace{2mm}
\noindent\textbf{WBF} Using ATSS bounding boxes showed the best score on the Acc@10 so it is chosen to be the base when fusing other results from detection models. As shown in Table ~\ref{table:wbf_result}, the detection performance increments as more boxes are merged together.

\vspace{2mm}
\noindent\textbf{Feature Concatenation}
We summarize performances of ensembled retrieval models using $WBF_{5}$ bounding boxes on the DeepFashion2 validation/test set in Table ~\ref{table:exp_retrieval_ensemble}.
By involving more models, we gain 0.019/0.014 increases on the Acc@10 results as compared to that of the single baseline retrieval model on DeepFashion2 validation/test set.

\subsubsection{What did not work well}

\vspace{2mm}
\noindent\textbf{PCA Whitening} In our experiments, dimensionality reduction by principal component analysis (PCA)~\cite{wold1987principal} always decreased performance by a small margin. Using the findings of our prior experiment, we applied whitening~\cite{sharif2014cnn} without dimensionality reduction, but that still did not add any improvements to the overall performance. Thus, we decided to use the full dimensions of the concatenated features to maximize the score.

\begin{table}
\begin{center}
{\small
\setlength\tabcolsep{3pt}
\begin{tabular}{|l|c|c|c|}
\hline
Ensemble Model & Validation (Acc@10) & Test (Acc@10)\\
\hline\hline
m4 (baseline) & 0.869435 & 0.840060 \\
\hline
m0-13 (w/o m6-9) & 0.875656 & 0.849722 \\
m0-13 (w/o m6-7) & 0.876141 & 0.853057 \\
m0-16 (w/o m7) & 0.876868 & 0.851860 \\
m1-16 (w/o m7) & 0.877192 & 0.851945 \\
m0-17 (w/o m7) & 0.876222 & 0.851945 \\
m1-16 & \textbf{0.888341} & \textbf{0.854168} \\
m1-17 & 0.888180 & 0.853356 \\
\hline
\end{tabular}
}
\end{center}
\vspace{-5mm}
\caption{Top-10 accuracy of retrieval ensembled models with ${WBF}_{5}$ bounding boxes on DeepFashion2 \textit{validation/test set}.}
\label{table:exp_retrieval_ensemble}
\vspace{-2mm}
\end{table}

\begin{table}
\begin{center}
\begin{adjustbox}{width=1.0\columnwidth,center}
\begin{tabular}{|l|c|c|c|c|}
\hline
Model & Detection & Search Method & Acc@1 & Acc@10 \\
\hline\hline
m4 & $WBF_5$ (590k) & NN & 0.668740 & 0.869435 \\
\hline
m4 & $WBF_5$ (126k) & NN & 0.650885 & 0.853761 \\
m4 & $WBF_5$ (126k) & NN + re-ranking & 0.661954 & 0.858124 \\
\hline
\end{tabular}
\end{adjustbox}
\end{center}
\vspace{-5mm}
\caption{The experimental results using k-reciprocal re-ranking on DeepFashion2 \textit{validation set}.}
\label{table:exp_k-reciprocal_encoding}
\vspace{-2mm}
\end{table}

\vspace{2mm}
\noindent\textbf{QE and DBA}
Query expansion (QE)~\cite{chum2007total} and database-side augmentation (DBA)~\cite{turcot2009better} replaces each feature point with a weighted sum of its top $k$ nearest neighbors and the point itself. The purpose of these techniques is to obtain rich and distinctive image representations by exploiting the nearest neighbors. However, QE decreased in overall performance while DBA gave a negligible increase. We speculate that this is because there were too many similar boxes proposed for the high recall query, hence the weighted sum of those boxes were not an informative image representation.

\vspace{2mm}
\noindent\textbf{Re-ranking}
In Table ~\ref{table:exp_k-reciprocal_encoding}, we present the experiment of k-reciprocal encoding~\cite{zhong2017re}, a well-known re-ranking technique in the Re-ID task, based on the DeepFashion2 validation set. This method requires much less bounding boxes to be present thus we used WBF to reduce the bounding boxes from 590k ($WBF_{5}$) to 126k ($WBF_{5}$). Interestingly, the k-reciprocal encoding helped increase the acc@10 by 0.02 but overall, the performance was still less than simply having reduced the boxes to 440k ($WBF_{5}$) using the WBF.

\begin{table}[]
\begin{center}
\begin{adjustbox}{width=0.9\columnwidth,center}
\begin{tabular}{|l|c|c|c|}
\hline
Year                  & Rank & Team                              & Acc@10            \\ \hline\hline
\multirow{3}{*}{2020} & 1    & Alibaba                           & 0.872082          \\  
                      & \textbf{2}    & \textbf{NAVER/LINE Vision (Ours)} & \textbf{0.854168} \\  
                      & 3    & DeepBlueAI                        & 0.848012          \\ \hline
\multirow{3}{*}{2019} & 1    & Hydra@ViSenze                     & 0.840658          \\  
                      & 2    & MM AI kakao                       & 0.823258          \\  
                      & 3    & DeepBlueAI                        & 0.816460          \\ \hline
\end{tabular}
\end{adjustbox}
\end{center}
\vspace{-5mm}
\caption{The final results of DeepFashion2 clothes retrieval challenge in 2020 and 2019.}
\label{table:result}
\vspace{-2mm}
\end{table}

%-------------------------------------------------------------------------
\section{Conclusion}
In this paper, we presented an effective pipeline for a real-world clothes retrieval system, including detection and retrieval task.
And also, we investigated various post-processing methods such as weighted boxes fusion, feature concatenation, and other techniques.
Table~\ref{table:result} shows the final result of the DeepFashion2 clothes retrieval challenge in 2020 and 2019.
With the proposed pipeline, we achieved 0.854168 on Acc@10 in the test phase and ranked on 2nd place in the DeepFashion2 clothes retrieval challenge 2020. 
Furthermore, we believe the considered pipeline and methods described in this paper can be generalized to fulfill visual search for images not only about fashion but also furniture, beauty products and toys.

%-------------------------------------------------------------------------

{\small
\bibliographystyle{ieee_fullname}
\bibliography{egbib}

\begin{thebibliography}{10}\itemsep=-1pt

\bibitem{Cai_2018_CVPR}
Zhaowei Cai and Nuno Vasconcelos.
\newblock Cascade r-cnn: Delving into high quality object detection.
\newblock In {\em The IEEE Conference on Computer Vision and Pattern
  Recognition (CVPR)}, June 2018.

\bibitem{chum2007total}
Ondrej Chum, James Philbin, Josef Sivic, Michael Isard, and Andrew Zisserman.
\newblock Total recall: Automatic query expansion with a generative feature
  model for object retrieval.
\newblock In {\em 2007 IEEE 11th International Conference on Computer Vision},
  pages 1--8. IEEE, 2007.

\bibitem{Dong_2019_ICCV}
Haoye Dong, Xiaodan Liang, Xiaohui Shen, Bochao Wang, Hanjiang Lai, Jia Zhu,
  Zhiting Hu, and Jian Yin.
\newblock Towards multi-pose guided virtual try-on network.
\newblock In {\em The IEEE International Conference on Computer Vision (ICCV)},
  October 2019.

\bibitem{DeepFashion2}
Yuying Ge, Ruimao Zhang, Lingyun Wu, Xiaogang Wang, Xiaoou Tang, and Ping Luo.
\newblock A versatile benchmark for detection, pose estimation, segmentation
  and re-identification of clothing images.
\newblock {\em CVPR}, 2019.

\bibitem{han2017viton}
Xintong Han, Zuxuan Wu, Zhe Wu, Ruichi Yu, and Larry~S Davis.
\newblock Viton: An image-based virtual try-on network.
\newblock In {\em CVPR}, 2018.

\bibitem{he2015deep}
Kaiming He, Xiangyu Zhang, Shaoqing Ren, and Jian Sun.
\newblock Deep residual learning for image recognition, 2015.

\bibitem{he2019bag}
Tong He, Zhi Zhang, Hang Zhang, Zhongyue Zhang, Junyuan Xie, and Mu Li.
\newblock Bag of tricks for image classification with convolutional neural
  networks.
\newblock In {\em Proceedings of the IEEE Conference on Computer Vision and
  Pattern Recognition}, pages 558--567, 2019.

\bibitem{hu2017squeezeandexcitation}
Jie Hu, Li Shen, Samuel Albanie, Gang Sun, and Enhua Wu.
\newblock Squeeze-and-excitation networks, 2017.

\bibitem{Huang_2015_ICCV}
Junshi Huang, Rogerio~S. Feris, Qiang Chen, and Shuicheng Yan.
\newblock Cross-domain image retrieval with a dual attribute-aware ranking
  network.
\newblock In {\em The IEEE International Conference on Computer Vision (ICCV)},
  December 2015.

\bibitem{1903.10663}
HeeJae Jun, ByungSoo Ko, Youngjoon Kim, Insik Kim, and Jongtack Kim.
\newblock Combination of multiple global descriptors for image retrieval, 2019.

\bibitem{Lin_2017_ICCV}
Tsung-Yi Lin, Priya Goyal, Ross Girshick, Kaiming He, and Piotr Dollar.
\newblock Focal loss for dense object detection.
\newblock In {\em The IEEE International Conference on Computer Vision (ICCV)},
  Oct 2017.

\bibitem{Liu2016MVCAD}
Kuan-Hsien Liu, Ting-Yen Chen, and Chu-Song Chen.
\newblock Mvc: A dataset for view-invariant clothing retrieval and attribute
  prediction.
\newblock In {\em ICMR '16}, 2016.

\bibitem{WhereToBuyItICCV15}
Svetlana Lazebnik Alexander C. Berg Tamara L.~Berg M.~Hadi~Kiapour, Xufeng~Han.
\newblock Where to buy it:matching street clothing photos in online shops.
\newblock In {\em International Conference on Computer Vision}, 2015.

\bibitem{radenovic2018fine}
Filip Radenovi{\'c}, Giorgos Tolias, and Ond{\v{r}}ej Chum.
\newblock Fine-tuning cnn image retrieval with no human annotation.
\newblock {\em IEEE transactions on pattern analysis and machine intelligence},
  41(7):1655--1668, 2018.

\bibitem{radenovi2017finetuning}
Filip Radenović, Giorgos Tolias, and Ondřej Chum.
\newblock Fine-tuning cnn image retrieval with no human annotation, 2017.

\bibitem{schroff2015facenet}
Florian Schroff, Dmitry Kalenichenko, and James Philbin.
\newblock Facenet: A unified embedding for face recognition and clustering.
\newblock In {\em Proceedings of the IEEE conference on computer vision and
  pattern recognition}, pages 815--823, 2015.

\bibitem{sharif2014cnn}
Ali Sharif~Razavian, Hossein Azizpour, Josephine Sullivan, and Stefan Carlsson.
\newblock Cnn features off-the-shelf: an astounding baseline for recognition.
\newblock In {\em Proceedings of the IEEE conference on computer vision and
  pattern recognition workshops}, pages 806--813, 2014.

\bibitem{1910.13302}
Roman Solovyev and Weimin Wang.
\newblock Weighted boxes fusion: ensembling boxes for object detection models,
  2019.

\bibitem{tolias2015particular}
Giorgos Tolias, Ronan Sicre, and Hervé Jégou.
\newblock Particular object retrieval with integral max-pooling of cnn
  activations, 2015.

\bibitem{turcot2009better}
Panu Turcot and David~G Lowe.
\newblock Better matching with fewer features: The selection of useful features
  in large database recognition problems.
\newblock In {\em 2009 IEEE 12th International Conference on Computer Vision
  Workshops, ICCV Workshops}, pages 2109--2116. IEEE, 2009.

\bibitem{wang2019multi}
Xun Wang, Xintong Han, Weilin Huang, Dengke Dong, and Matthew~R Scott.
\newblock Multi-similarity loss with general pair weighting for deep metric
  learning.
\newblock In {\em Proceedings of the IEEE Conference on Computer Vision and
  Pattern Recognition}, pages 5022--5030, 2019.

\bibitem{wen2016discriminative}
Yandong Wen, Kaipeng Zhang, Zhifeng Li, and Yu Qiao.
\newblock A discriminative feature learning approach for deep face recognition.
\newblock In {\em European conference on computer vision}, pages 499--515.
  Springer, 2016.

\bibitem{wold1987principal}
Svante Wold, Kim Esbensen, and Paul Geladi.
\newblock Principal component analysis.
\newblock {\em Chemometrics and intelligent laboratory systems}, 2(1-3):37--52,
  1987.

\bibitem{zhang2020bridging}
Shifeng Zhang, Cheng Chi, Yongqiang Yao, Zhen Lei, and Stan~Z. Li.
\newblock Bridging the gap between anchor-based and anchor-free detection via
  adaptive training sample selection.
\newblock In {\em CVPR}, 2020.

\bibitem{zhong2017re}
Zhun Zhong, Liang Zheng, Donglin Cao, and Shaozi Li.
\newblock Re-ranking person re-identification with k-reciprocal encoding.
\newblock In {\em Proceedings of the IEEE Conference on Computer Vision and
  Pattern Recognition}, pages 1318--1327, 2017.

\bibitem{1904.07850}
Xingyi Zhou, Dequan Wang, and Philipp Krähenbühl.
\newblock Objects as points, 2019.

\end{thebibliography}
}

\end{document}